\documentclass[twocolumn]{article}
\usepackage[top=1in, bottom=1in, left=0.75in, right=0.75in]{geometry}

\usepackage{graphicx}
\usepackage{amsmath}
\usepackage{amssymb}
\usepackage{bm}
\usepackage{url}
\usepackage{wrapfig}
\usepackage[usenames,dvipsnames,svgnames,table]{xcolor}
\usepackage{soul}	
\usepackage{authblk} 
\usepackage{dsfont} 
\usepackage{fancyvrb}
\newtheorem{theorem}{Theorem}

\newif\ifcomment
\commenttrue      

\ifcomment
\newcommand{\qirong}[1]{{\color{blue}{\bf\sf [Qirong: #1]}}} 

\newcommand{\daiwei}[1]{{\color{purple}{\bf\sf [Daiwei: #1]}}}

\newcommand{\jinliang}[1]{{\color{red}{\bf\sf [Jinliang: #1]}}}

\else

\newcommand{\qirong}[1]{{\color{blue}{}}}
\newcommand{\abhi}[1]{{\color{magenta}{}}}
\newcommand{\daiwei}[1]{{\color{purple}{}}}
\newcommand{\jinliang}[1]{{\color{green}{}}}
\fi

\newcommand{\bx} {\textbf{x}}

\newcommand{\by} {\textbf{y}}

\newcommand{\bu} {\textbf{u}}

\newcommand{\bbeta}{\boldsymbol{\beta}}

\newcommand{\bX}{\mathbf{X}}

\newcommand{\gvec}{\mathbf{g}}
\newcommand{\xvec}{\mathbf{x}}
\newcommand{\wvec}{\mathbf{w}}
\newcommand{\zvec}{\mathbf{z}}
\newcommand{\RR}{\mathds{R}}
\newcommand{\EE}{\mathds{E}}

\begin{document}
\title{Petuum: A New Platform for Distributed Machine Learning on Big Data}

\author[1]{\rm Eric P. Xing}
\author[2]{\rm Qirong Ho}
\author[1]{\rm Wei Dai}
\author[1]{\rm Jin Kyu Kim}
\author[1]{\rm Jinliang Wei}
\author[1]{\rm Seunghak Lee}
\author[1]{\rm Xun Zheng}
\author[1]{\rm Pengtao Xie}
\author[1]{\rm Abhimanu Kumar}
\author[1]{\rm Yaoliang Yu}
\affil[1]{School of Computer Science, Carnegie Mellon University}
\affil[2]{Institute for Infocomm Research, A*STAR, Singapore}
\affil[ ]{\textit {\{epxing,wdai,jinkyuk,jinlianw,seunghak,xunzheng,pengtaox,yaoliang\}@cs.cmu.edu}}
\affil[ ]{\textit {\{hoqirong,abhimanyu.kumar\}@gmail.com}}

\maketitle

\vspace{-0.5cm}
\begin{abstract}
What is a systematic way to efficiently apply a wide spectrum of advanced ML programs to industrial scale problems, using Big Models (up to 100s of billions of parameters) on Big Data (up to terabytes or petabytes)? Modern parallelization strategies employ fine-grained operations and scheduling beyond the classic bulk-synchronous processing paradigm popularized by MapReduce, or even specialized graph-based execution that relies on graph representations of ML programs. The variety of approaches tends to pull systems and algorithms design in different directions, and it remains difficult to find a universal platform applicable to a wide range of ML programs at scale. We propose a general-purpose framework that systematically addresses data- and model-parallel challenges in large-scale ML, by observing that many ML programs are fundamentally optimization-centric and admit error-tolerant, iterative-convergent algorithmic solutions. This presents unique opportunities for an integrative system design, such as bounded-error network synchronization and dynamic scheduling based on ML program structure. We demonstrate the efficacy of these system designs versus well-known implementations of modern ML algorithms, allowing ML programs to run in much less time and at considerably larger model sizes, even on modestly-sized compute clusters.\\[-0.6cm]
\end{abstract}

\vspace{-0.3cm}
\section{Introduction}
\vspace{-0.1cm}

Machine learning (ML) is becoming a primary mechanism for extracting information from data. However, the surging volume of Big Data from Internet activities and sensory advancements, and the increasing needs for Big Models for ultra high-dimensional problems have put tremendous pressure on ML methods to scale beyond a single machine, due to both space and time bottlenecks. 
For example, the Clueweb 2012 web crawl\footnote{http://www.lemurproject.org/clueweb12.php/} contains over 700 million web pages as 27TB of text data, while photo-sharing sites such as Flickr, Instagram and Facebook are anecdotally known to possess 10s of billions of images, again taking up TBs of storage. It is highly inefficient, if possible, to use such big data sequentially in a batch or scholastic fashion in a typical iterative ML algorithm. 
On the other hand, state-of-the-art image recognition systems have now embraced large-scale deep learning models with billions of parameters~\cite{cat_recognizer12}; 
topic models with up to $10^6$ topics can cover long-tail semantic word sets for substantially improved online advertising~\cite{peacock,yuan2015lightlda}; and very-high-rank matrix factorization yields improved prediction on collaborative filtering problems~\cite{large_netflix08}. Training such big models with a single machine can be prohibitively slow, if possible. 


Despite the recent rapid development of many new ML models and algorithms aiming at scalable application
\cite{dean2012large,williamson2013parallel,hoffman2013stochastic,slow_learners,agarwal_delay,shotgun}, 
adoption of these technologies remains generally unseen in the wider data mining, NLP, vision, and other application communities for big problems, especially those built on advanced probabilistic or optimization programs. We suggest that, from the scalable execution point of view, what prevents many state-of-the-art ML models and algorithms from being more widely applied at Big-Learning scales is the difficult migration from an academic implementation, often specialized for a small, well-controlled computer platform such as desktop PCs and small lab-clusters, to a big, less predictable platform such as a corporate cluster or the cloud, where correct execution of the original programs require careful control and mastery of low-level details of the distributed environment and resources through highly nontrivial distributed programming. 

Many platforms have provided partial solutions to bridge this research-to-production gap: 
while Hadoop~\cite{white2012hadoop} is a popular and easy to program platform, the simplicity of its MapReduce abstraction makes it difficult to exploit ML properties such as error tolerance (at least, not without considerable engineering effort to bypass MapReduce limitations), and its performance on many ML programs has been surpassed by alternatives~\cite{spark,graphlab12}. One such alternative is Spark~\cite{spark}, which generalizes MapReduce and scales well on data while offering an accessible programming interface; yet, Spark does not offer fine-grained scheduling of computation and communication, which has been shown to be hugely advantageous, if not outright necessary, for fast and correct execution of advanced ML algorithms~\cite{dai2015high}. Graph-centric platforms such as GraphLab~\cite{graphlab12} and Pregel~\cite{pregel} efficiently partition graph-based models with built-in scheduling and consistency mechanisms; but ML programs such as topic modeling and regression either do not admit obvious graph representations, or a graph representation may not be the most efficient choice; moreover, due to limited theoretical work, it is unclear whether asynchronous graph-based consistency models and scheduling will always yield correct execution of such ML programs. Other systems provide low-level programming interfaces~\cite{piccolo10, muli_osdi14}, that, while powerful and versatile, do not yet offer higher-level general-purpose building blocks such as scheduling, model partitioning strategies, and managed communication that are key to simplifying the adoption of a wide range of ML methods. In summary, existing systems supporting distributed ML each manifest a unique tradeoff on efficiency, correctness, programmability, and generality.

\begin{figure}[t]
\centering
\includegraphics[width=1\linewidth]{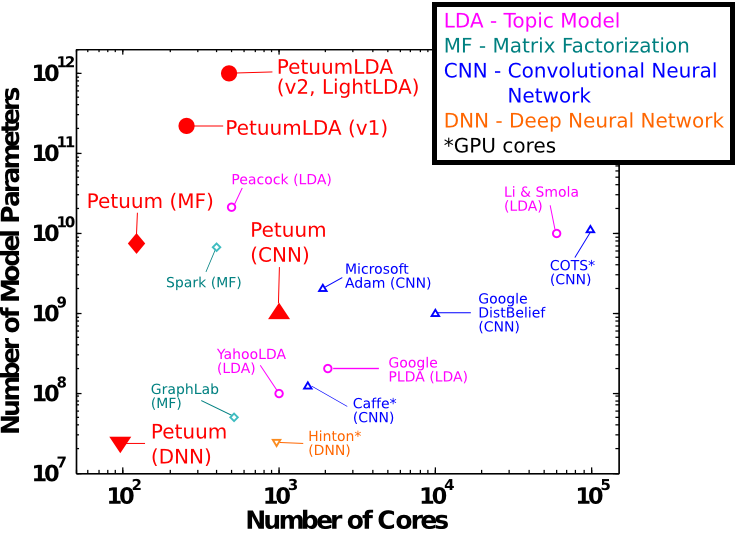}
\vspace{-0.5cm}
\caption{\small The scale of Big ML efforts in recent literature. A key goal of Petuum is to enable larger ML models to be run on fewer resources, even relative to
highly-specialized implementations.}
\label{fig:bigml_scale}
\vspace{-0.4cm}
\end{figure}

In this paper, we explore the problem of building a distributed machine learning framework with a new angle toward the efficiency, correctness, programmability, and generality tradeoff.  We observe that, a hallmark of most (if not all) ML programs is that they are defined by an explicit objective function over data (e.g., likelihood, error-loss, graph cut),
and the goal is to attain optimality of this function, in the space defined by the model parameters and other intermediate variables.
Moreover, these algorithms all bear a common style, in that they resort to an iterative-convergent procedure (see Eq.~\ref{eq:master_general}). It is noteworthy that iterative-convergent computing tasks are vastly different from conventional programmatic 
computing tasks (such as database queries and keyword extraction), which reach correct solutions only if every deterministic operation is correctly executed, and strong consistency is guaranteed on the intermediate program state --- thus, operational objectives such as fault tolerance and strong consistency are absolutely necessary. However, an ML program's true goal is fast, efficient convergence to an optimal solution, and we argue that fine-grained fault tolerance and strong consistency are but one vehicle to achieve this goal, and might not even be the most efficient one.

We present a new distributed ML framework, {\it Petuum},
built on an ML-centric optimization-theoretic principle, as opposed to various operational objectives explored earlier. We begin by formalizing ML algorithms as {\it iterative-convergent} programs, which encompass a large space of modern ML such as stochastic gradient descent, MCMC for determining point estimates in latent variable models~\cite{griffiths2004finding}, coordinate descent, variational methods for graphical models~\cite{hoffman2013stochastic}, proximal optimization for structured sparsity problems~\cite{xichen:UAI:11}, among others. To our knowledge, no existing ML platform has considered such a wide spectrum of ML algorithms, which exhibit diverse representation abstractions, model and data access patterns, and synchronization and scheduling requirements. So what are the shared properties across such a ``zoo of ML algorithms''? We believe that the key lies in the recognition of a clear dichotomy between {\it data} (which is conditionally independent
and persistent throughout the algorithm) and {\it model} (which is internally coupled, and is transient before converging to an optimum). This inspires a simple yet statistically-rooted bimodal approach to parallelism: {\it data parallel} and {\it model parallel} distribution and execution of a big ML program over a cluster of machines. This {\it data parallel}, {\it model parallel} approach keenly exploits the unique statistical nature of ML algorithms, particularly the following three properties: (1) Error tolerance --- iterative-convergent algorithms are often robust against limited errors in intermediate calculations; (2) Dynamic structural dependency --- during execution, the changing correlation strengths between model parameters are critical to efficient parallelization; (3) Non-uniform convergence --- the number of steps required for a parameter to converge can be highly skewed across parameters.
The core goal of Petuum is to execute these iterative updates in a manner that quickly converges to an optimum of the ML program's objective function, by exploiting these three statistical properties of ML, which we argue are fundamental to efficient large-scale ML in cluster environments.

This design principle contrasts that of several existing frameworks discussed earlier. 
For example, central to the Spark framework~\cite{spark} is the principle of perfect fault tolerance and recovery, supported by a persistent memory architecture (Resilient Distributed Datasets); whereas central to the GraphLab framework is the principle of local and global consistency, supported by a vertex programming model (the Gather-Apply-Scatter abstraction). While these design principles reflect important aspects of correct ML algorithm execution --- e.g., atomic recoverability of each computing step (Spark), or consistency satisfaction for all subsets of model variables (GraphLab) --- 
some other important aspects, such as the three statistical properties discussed above, or perhaps ones that could be more fundamental and general, and which could open more room for efficient system designs, remain unexplored.  

To exploit these properties, Petuum introduces three novel system objectives grounded in the aforementioned key properties of ML programs, in order to accelerate their convergence at scale: (1) Petuum synchronizes the parameter states with a bounded staleness guarantee, which achieves provably correct outcomes due to the error-tolerant nature of ML, but at a much cheaper communication cost than conventional per-iteration bulk synchronization; (2) Petuum offers dynamic scheduling policies that take into account the changing structural dependencies between model parameters, so as to minimize parallelization error and synchronization costs; and (3) Since parameters in ML programs exhibit non-uniform convergence costs (i.e. different numbers of updates required), Petuum prioritizes computation towards non-converged model parameters, so as to achieve faster convergence.

To demonstrate this approach, we show how a data-parallel and a model-parallel algorithm can be implemented on Petuum, allowing them to scale to large model sizes with improved algorithm convergence times. This is illustrated in Figure~\ref{fig:bigml_scale}, where Petuum is able to solve a range of ML problems at reasonably large model scales, even on relatively modest clusters (10-100 machines) that are within reach of most ML practitioners. The experiments section provides more detailed benchmarks on a range of ML programs: topic modeling, matrix factorization, deep learning, Lasso regression, and distance metric learning. These algorithms are only a subset of the full open-source Petuum ML library\footnote{Petuum is available as open source at \url{http://petuum.org}.}, which includes more algorithms not explored in this paper: random forests, K-means, sparse coding, MedLDA, SVM, multi-class logistic regression, with many others being actively developed for future releases.

\vspace{-0.3cm}
\section{Preliminaries: On Data and \\ Model Parallelism}
\vspace{-0.1cm}

We begin with a principled formulation of iterative-convergent ML programs, which exposes a dichotomy of data and model, that inspires the parallel system architecture (\S 3), algorithm design (\S 4), and theoretical analysis (\S5) of Petuum.   
Consider the following programmatic view of ML as iterative-convergent programs, driven by an objective function: 

\noindent 
{\bf Iterative-Convergent ML Algorithm: } Given data $D$ and model $\mathcal{L}$ (i.e., a fitness function such as RMS loss, likelihood, margin), a typical ML problem can be grounded as executing the following update equation iteratively, until the model state (i.e., parameters and/or latent variables) $A$ reaches some stopping criteria:
\begin{align}
A^{(t)} = F(A^{(t-1)}, \Delta_{\mathcal{L}}(A^{(t-1)},D))
\label{eq:master_general}
\end{align}
where superscript $(t)$ denotes iteration. The update function $\Delta_{\mathcal{L}}()$ (which improves the loss $\mathcal{L}$) performs computation on data $D$ and model state $A$, and outputs intermediate results to be aggregated by $F()$. For simplicity, in the rest of the paper we omit $\mathcal{L}$ in the subscript with the understanding that all ML programs of our interest here bear an explicit loss function that can be used to monitor the quality of convergence and solution, as oppose to heuristics or procedures not associated such a loss function. 

\begin{figure}[t]
\vspace{-0.6cm}
\centering
\includegraphics[width=0.7\linewidth]{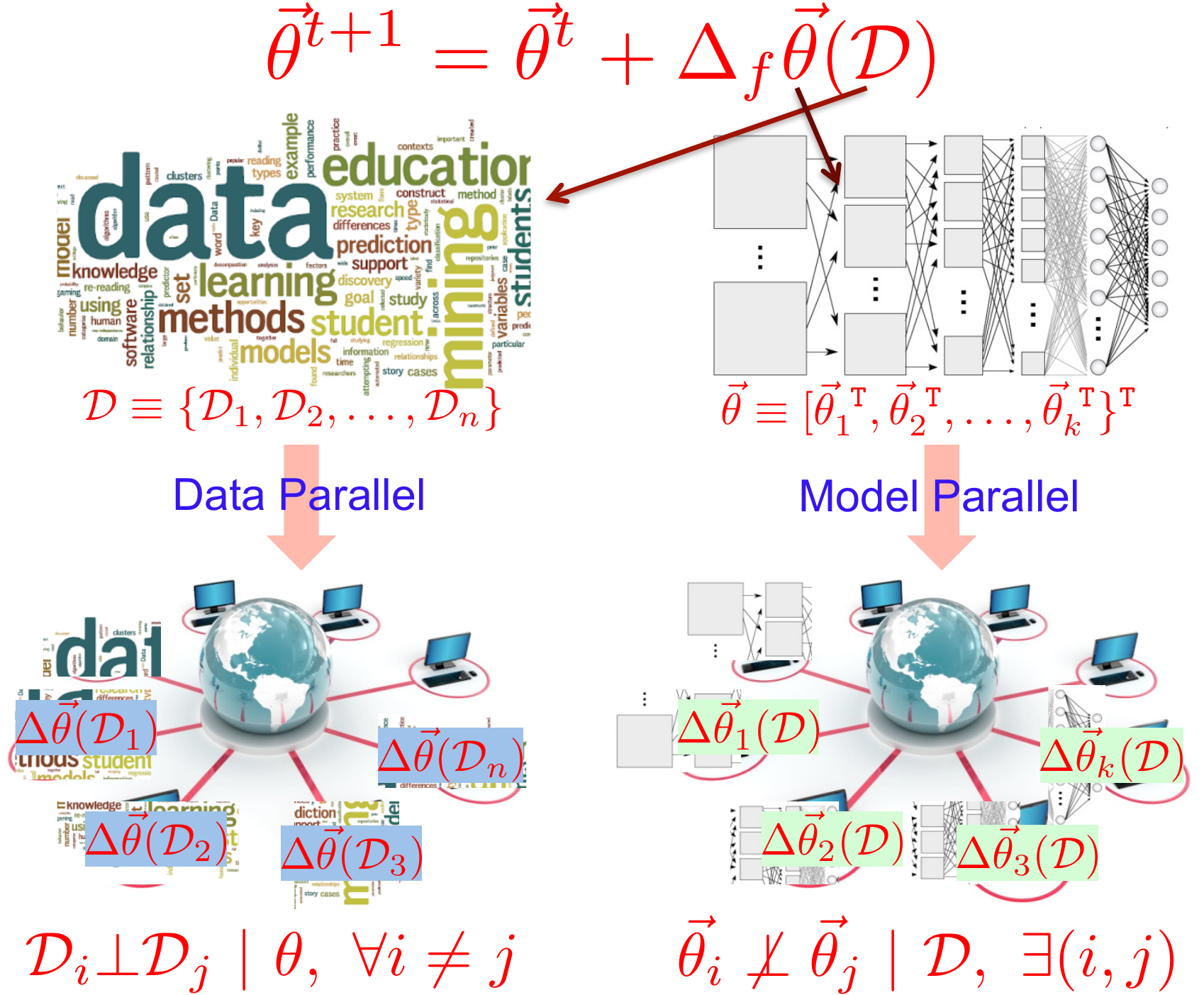}
\vspace{-0.4cm}
\caption{\small The difference between data and model parallelism: data samples are always conditionally independent given the model, but there are some model parameters that are not independent of each other.}
\label{fig:data_model_parallel_difference}
\vspace{-0.4cm}
\end{figure}

In large-scale ML, both data $D$ and model $A$ can be very large. {\it Data-parallelism}, in which data is divided across machines, is a common strategy for solving Big Data problems, while {\it model-parallelism}, which divides the ML model, is common for Big Models.
Below, we discuss the (different) mathematical implications of each parallelism (see Fig.~\ref{fig:data_model_parallel_difference}). 

\vspace{-0.2cm}
\subsection{Data Parallelism}

In {\it data-parallel} ML, the data $D$ is partitioned and assigned to computational workers (indexed by $p=1..P$); we denote the $p$-th data partition by $D_p$. We assume that the function $\Delta()$ can be applied to each of these data subsets independently, yielding a data-parallel update equation:
\begin{equation}
\textstyle A^{(t)} = F(A^{(t-1)}, \sum_{p=1}^{P} \Delta(A^{(t-1)},D_p) ).
\label{eq:sspGeneralF}
\end{equation}
In this definition, we assume that the $\Delta()$ outputs are aggregated via summation, which is commonly seen in stochastic gradient descent or sampling-based algorithms.
For example, in distance metric learning problem which is optimized with stochastic gradient descent (SGD), the data pairs are partitioned over different workers, and the intermediate results (subgradients) are computed on each partition and are summed before applied to update the model parameters.
Other algorithms can also be expressed in this form, 
such as variational EM algorithms $A^{(t)} = \sum_{p=1}^{P} \Delta(A^{(t-1)},D_p)$. Importantly, this {\it additive updates} property allows the updates $\Delta()$ to be aggregated at each local worker before transmission over the network, which is crucial because CPUs can produce updates $\Delta()$ much faster than they can be (individually) transmitted over the network. Additive updates are the foundation for a host of techniques to speed up data-parallel execution, such as minibatch, asynchronous and bounded-asynchronous execution, and parameter servers.
Key to the validity of additivity of updates from different workers is the notion of {\it independent and identically distributed (iid)} data, which is assumed for many ML programs, and implies that each parallel worker contributes ``equally" (in a statistical sense) to the ML algorithm's progress via $\Delta()$, no matter which data subset $D_p$ it uses.

\vspace{-0.2cm}
\subsection{Model Parallelism}




In {\it model-parallel} ML, the model $A$ is partitioned and assigned to workers $p=1..P$ and updated therein in parallel, running update functions $\Delta()$. Unlike data-parallelism, each update function $\Delta()$ also takes a scheduling function $S_p^{(t-1)}()$,
which restricts $\Delta()$ to operate on a subset of the model parameters $A$:
\begin{equation}
A^{(t)} = F\left(A^{(t-1)}, \{ \Delta(A^{(t-1)}, S_p^{(t-1)}(A^{(t-1)})) \}_{p=1}^P \right),
\label{eq:modelpara}
\end{equation}
where we have omitted the data $D$ for brevity and clarity. $S_p^{(t-1)}()$ outputs a set of indices $\{j_1,j_2,\dots,\}$,
so that $\Delta()$ only performs updates on $A_{j_1},A_{j_2},\dots$ --- we refer to such selection of model parameters as {\it scheduling}.

Unlike data-parallelism which enjoys iid data properties, the model parameters $A_j$ are not, in general, independent of each other (Figure \ref{fig:data_model_parallel_difference}), and it has been established that model-parallel algorithms can only be effective if the parallel updates are restricted to independent (or weakly-correlated) parameters~\cite{lee2014model,shotgun,feature_cluster,graphlab12}. Hence, our definition of model-parallelism includes a {\it global scheduling mechanism} that can select carefully-chosen parameters for parallel updating.

The scheduling function $S()$ opens up a large design space, such as fixed, randomized, or even dynamically-changing scheduling on the whole space, or a subset of, the model parameters. $S()$ not only can provide {\it safety and correctness} (e.g., by selecting independent parameters and thus minimize parallelization error), but can offer substantial {\it speed-up} (e.g., by prioritizing computation onto non-converged parameters). In the Lasso example, Petuum uses $S()$ to select coefficients that are weakly correlated (thus preventing divergence), while at the same time prioritizing coefficients far from zero (which are more likely to be non-converged).

\vspace{-0.2cm}
\subsection{Implementing Data- \\ and Model-Parallel Programs}


Data- and model-parallel programs are {\it stateful}, in that they continually update shared model parameters $A$. Thus, an ML platform needs to synchronize $A$ across all running threads and processes, and this should be done in a high-performance non-blocking manner that still guarantees convergence. Ideally, the platform should also offer easy, global-variable-like access to $A$ (as opposed to cumbersome message-passing, or non-stateful MapReduce-like functional interfaces). If the program is model-parallel, it may require fine control over parameter scheduling to avoid non-convergence; such capability is not available in Hadoop, Spark nor GraphLab without code modification. Hence, there is an opportunity to address these considerations via a platform tailored to data- and model-parallel ML.

\vspace{-0.3cm}
\section{Petuum -- \\ a Platform for Distributed ML}
\vspace{-0.1cm}

A core goal of Petuum is to allow practitioners to easily implement data-parallel and model-parallel ML algorithms.
Petuum
provides APIs to key systems that make data- and model-parallel programming easier: (1) a {\it parameter server} system, which allows programmers to access global model state $A$ from any machine via a convenient {\it distributed shared-memory} interface that resembles single-machine programming, and adopts a bounded-asychronous consistency model that preserves data-parallel convergence guarantees, thus freeing users from explicit network synchronization; (2) a {\it scheduler}, which allows fine-grained control over the parallel ordering of model-parallel updates $\Delta()$ --- in essence, the scheduler allows users to define their own ML application consistency rules.

\vspace{-0.2cm}
\subsection{Petuum System Design}

\begin{figure}[t]
\centering
\includegraphics[width=1.0\linewidth]{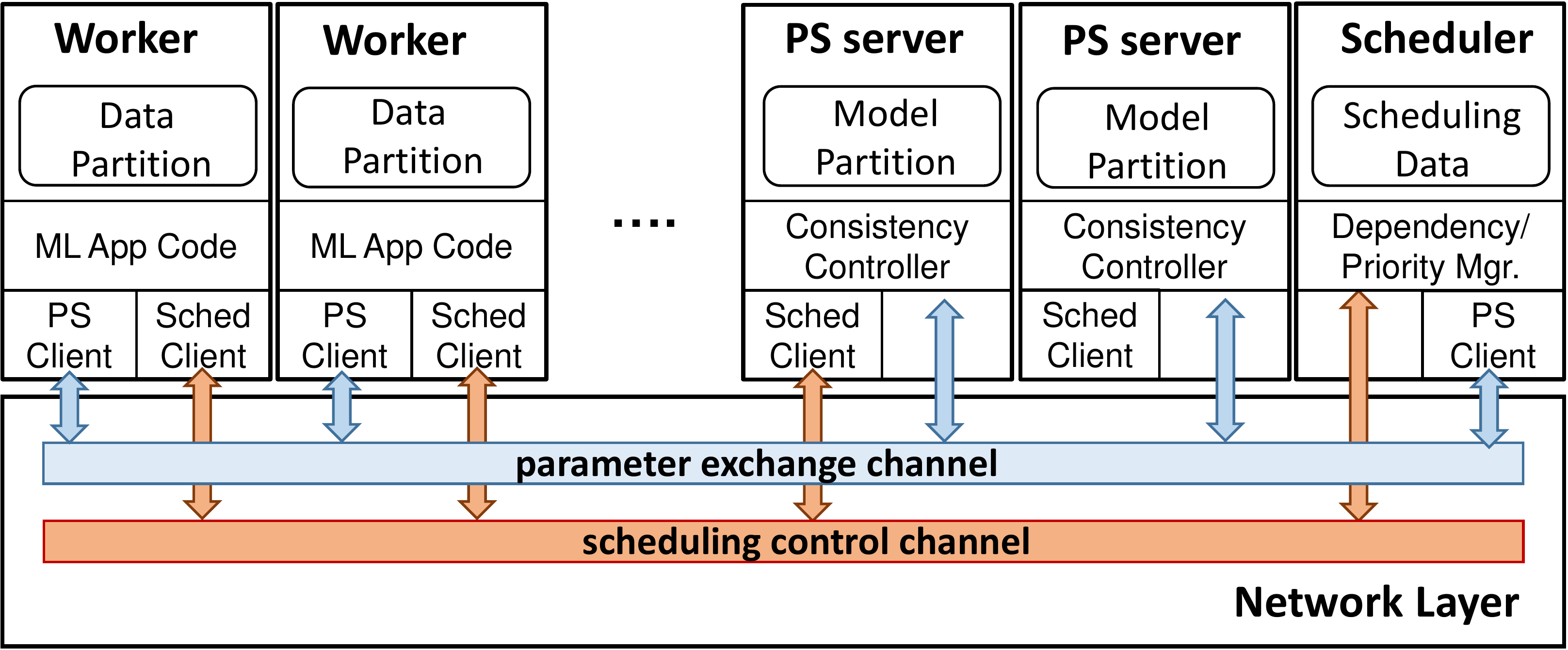}
\vspace{-0.6cm}
\caption{\small Petuum system: scheduler, workers, parameter servers.}
\label{fig:stack}
\vspace{-0.4cm}
\end{figure}

ML algorithms exhibit several principles that can be exploited to speed up distributed ML programs: dependency structures between parameters, non-uniform convergence of parameters, and a limited degree of error tolerance
\cite{ho13, dai2015high, lee2014model, priter, muli_osdi14, graphlab12}.
Petuum allows practitioners to write data-parallel and model-parallel ML programs that exploit these principles,
and can be scaled to Big Data and Big Model applications.
The Petuum system comprises three components (Fig.~\ref{fig:stack}): scheduler, workers, and parameter server, and Petuum ML programs are written in C++ (with Java support coming in the near future).


\noindent
{\bf Scheduler: }
The scheduler 
system enables model-parallelism, by allowing users to control which model parameters are updated by worker machines. This is performed through a user-defined scheduling function {\tt schedule()} (corresponding to $S_p^{(t-1)}()$), which outputs a set of parameters for each worker --- for example, a simple schedule might pick a random parameter for every worker, while a more complex scheduler (as we will show) may pick parameters according to multiple criteria, such as pair-wise independence or distance from convergence.
The scheduler sends the {\it identities} of these parameters
to workers via the scheduling control channel (Fig.~\ref{fig:stack}), while the actual parameter values are delivered through a parameter server system that we will soon explain; the scheduler is responsible only for deciding {\it which} parameters to update. Later, we will discuss some of the theoretical guarantees enjoyed by model-parallel schedules.

Several common patterns for schedule design are worth highlighting: the simplest option is a \textbf{fixed-schedule} ({\tt schedule\_fix()}), which dispatches model parameters $A$ in a pre-determined order (as is common in existing ML algorithm implementations).
Static, round-robin schedules (e.g. repeatedly loop over all parameters) fit the {\tt schedule\_fix()} model. Another type of schedule is \textbf{dependency-aware} ({\tt schedule\_dep()}) scheduling, which allows re-ordering of variable/parameter updates to accelerate model-parallel ML algorithms such as Lasso regression. This type of schedule analyzes the dependency structure over model parameters $A$, in order to determine their best parallel execution order.
Finally, \textbf{prioritized scheduling ({\tt schedule\_pri()})} exploits uneven convergence in ML, by prioritizing subsets of variables $U^{sub}\subset A$ according to algorithm-specific criteria, such as the magnitude of each parameter, or boundary conditions such as KKT.

Because scheduling functions {\tt schedule()} may be compute-intensive, Petuum uses pipelining to overlap scheduling computations {\tt schedule()} with worker execution, so workers are always doing useful computation. In addition, the scheduler is responsible for central aggregation via the {\tt pull()} function (corresponding to $F()$), if it is needed.


\noindent
{\bf Workers: }
Each worker $p$ receives parameters to be updated from the scheduler function {\tt schedule()}, and then runs parallel update functions {\tt push()} (corresponding to $\Delta()$) on data $D$. Petuum intentionally does not specify a data abstraction, so that any data storage system may be used --- workers may read from data loaded into memory, or from disk, or over a distributed file system or database such as HDFS. Furthermore, workers may touch the data in any order desired by the programmer: in data-parallel stochastic algorithms, workers might sample one data point at a time, while in batch algorithms, workers might instead pass through all data points in one iteration.
While {\tt push()} is being executed, the model state $A$ is automatically synchronized with the parameter server via the parameter exchange channel, using a distributed shared memory programming interface that conveniently resembles single-machine programming. After the workers finish {\tt push()}, the scheduler may use the new model state to generate future scheduling decisions.

\noindent
{\bf Parameter Server: } 
The parameter server (PS) provides global access to model parameters $A$, via a convenient distributed shared memory API that is similar to table-based or key-value stores. To take advantage of ML-algorithmic principles, the PS implements the Stale Synchronous Parallel (SSP) consistency model~\cite{ho13,dai2015high}, which reduces network synchronization and communication costs, while maintaining bounded-staleness convergence guarantees implied by SSP. We will discuss these guarantees in more detail later.

\vspace{-0.2cm}
\subsection{Programming Interface}

\begin{figure}[t]
{\scriptsize
\begin{Verbatim}[frame=single]
// Petuum Program Structure
\end{Verbatim}
\vspace{-0.7cm}
\begin{Verbatim}[frame=single]
schedule() {
  // This is the (optional) scheduling function
  // It is executed on the scheduler machines
  A_local = PS.get(A)  // Parameter server read
  PS.inc(A,change)  // Can write to PS here if needed
  // Choose variables for push() and return
  svars = my_scheduling(DATA,A_local)
  return svars
}
\end{Verbatim}
\vspace{-0.7cm}
\begin{Verbatim}[frame=single]
push(p = worker_id(), svars = schedule()) {
  // This is the parallel update function
  // It is executed on each of P worker machines
  A_local = PS.get(A)  // Parameter server read
  // Perform computation and send return values to pull()
  // Or just write directly to PS
  change1 = my_update1(DATA,p,A_local)
  change2 = my_update2(DATA,p,A_local)
  PS.inc(A,change1)  // Parameter server increment
  return change2
}
\end{Verbatim}
\vspace{-0.7cm}
\begin{Verbatim}[frame=single]
pull(svars = schedule(), updates = (push(1), ..., push(P)) ) {
  // This is the (optional) aggregation function
  // It is executed on the scheduler machines
  A_local = PS.get(A)  // Parameter server read
  // Aggregate updates from push(1..P) and write to PS
  my_aggregate(A_local,updates)
  PS.put(A,change)  // Parameter server overwrite
}
\end{Verbatim}
}
\vspace{-0.7cm}
\caption{\small Petuum Program Structure.}
\label{fig:petuum_interface}
\vspace{-0.5cm}
\end{figure}

Figure \ref{fig:petuum_interface} shows a basic Petuum program, consisting of a central scheduler function {\tt schedule()}, a parallel update function {\tt push()}, and a central aggregation function {\tt pull()}. The model variables $A$ are held in the parameter server, which can be accessed at any time from any function via the {\tt PS} object. The {\tt PS} object can be accessed from any function, and has 3 functions: {\tt PS.get()} to read a parameter, {\tt PS.inc()} to add to a parameter, and {\tt PS.put()} to overwrite a parameter. With just these operations, the SSP consistency model automatically ensures parameter consistency between all Petuum components; no additional user programming is necessary. Finally, we use {\tt DATA} to represent the data $D$; as noted earlier, this can be any 3rd-party data structure, database, or distributed file system. 

\vspace{-0.2cm}
\section{Petuum Parallel Algorithms}

Now we turn to development of parallel algorithms for large-scale distributed ML problems, in light of the data and model parallel principles underlying Petuum. We focus on a new data-parallel Distance Metric Learning algorithm, and a new model-parallel Lasso algorithm, but our strategies apply to a broad spectrum of other ML problems as briefly discussed at the end of this section. We show that with the Petuum system framework, we can easily realize these algorithms on distributed clusters without dwelling on low level system programming, or non-trivial recasting of our ML problems into  representations such as RDDs or vertex programs. Instead our ML problems can be coded at a high level, more akin to Matlab or R.

\subsection{Data-Parallel Distance Metric Learning}

Let us first consider a large-scale 
Distance Metric Learning (DML) problem. 
DML improves the performance of other ML programs such as clustering, by allowing domain experts to incorporate prior knowledge of the form ``data points $x$, $y$ are similar (or dissimilar)"~\cite{xing2002distance} --- for example, we could enforce that ``books about science are different from books about art". The output is a distance function $d(x,y)$ that captures the aforementioned prior knowledge. Learning a proper distance metric~\cite{davis2007information,xing2002distance} is essential for many distance based data mining and machine learning algorithms, such as retrieval, k-means clustering and k-nearest neighbor (k-NN) classification. DML has not received much attention in the Big Data setting, and we are not aware of any distributed implementations of DML.


The most popular version of DML tries to learn a Mahalanobis distance matrix $M$ (symmetric and positive-semidefinite), which can then be used to measure the distance between two samples $D(x,y) = (x-y)^{\mathsf{T}}M(x-y)$.
Given a set of ``similar" sample pairs $\mathcal{S}=\{(x_{i},y_{i})\}_{i=1}^{|\mathcal{S}|}$, and a set of ``dissimilar" pairs $\mathcal{D}=\{(x_{i},y_{i})\}_{i=1}^{|\mathcal{D}|}$, DML learns the Mahalanobis distance by optimizing
\begin{equation}
\label{eq:dml_opt_1}
\begin{array}{ll}
\textrm{min}_{M}&\sum\limits_{(x,y)\in \mathcal{S}}(x-y)^{\mathsf{T}}M(x-y)\\
s.t.&(x-y)^{\mathsf{T}}M(x-y)\geq1, \forall (x,y)\in\mathcal{D}\\
&M\succeq 0
\end{array}
\end{equation}
where $M\succeq 0$ denotes that $M$ is required to be positive semidefinite. This optimization problem tries to minimize the Mahalanobis distances between all pairs labeled as similar while separating dissimilar pairs with a margin of 1.

In its original form, this optimization problem is difficult to parallelize due to the constraint set. To create a data-parallel optimization algorithm and implement it on Petuum, we shall relax the constraints via slack variables (similar to SVMs). First, we replace $M$ with $L^{\mathsf{T}}L$,
and introduce slack variables $\xi$ to relax the hard constraint in Eq.(\ref{eq:dml_opt_1}), yielding
\begin{equation}
\label{eq:dml_opt_3}
\begin{array}{ll}
\textrm{min}_{L}&\sum\limits_{(x,y)\in \mathcal{S}}\|L(x-y)\|^{2}+\lambda \sum\limits_{(x,y)\in \mathcal{D}}\xi_{x,y}\\
s.t.&\|L(x-y)\|^{2}\geq1-\xi_{x,y}, \xi_{x,y}\geq 0, \forall (x,y)\in\mathcal{D}\\
\end{array}
\end{equation}
Using hinge loss, the constraint in Eq.(\ref{eq:dml_opt_3}) can be eliminated, yielding an unconstrained optimization problem:
\begin{equation}
\label{eq:dml_opt_4}
\begin{array}{ll}
\textrm{min}_{L}&\sum\limits_{(x,y)\in \mathcal{S}}\|L(x-y)\|^{2}\\
&+\lambda \sum\limits_{(x,y)\in \mathcal{D}}\textrm{max}(0, 1-\|L(x-y)\|^{2})\\
\end{array}
\end{equation}
Unlike the original constrained DML problem, this relaxation is fully data-parallel, because it now treats the dissimilar pairs as iid data to the loss function (just like the similar pairs); hence, it can be solved via data-parallel Stochastic Gradient Descent (SGD).
SGD can be naturally parallelized over data, and we partition the data pairs onto $P$ machines. Every iteration, each machine $p$ randomly samples a minibatch of similar pairs $\mathcal{S}_{p}$ and dissimilar pairs $\mathcal{D}_{p}$ from its data shard, and computes the following update to $L$:
\begin{equation}
\label{eq:dml_grad}
\begin{array}{lll}
\bigtriangleup L_{p}&=& \sum_{(x,y)\in \mathcal{S}_{p}}2L(x-y)(x-y)^{\mathsf{T}}\\
&-& \sum_{(a,b)\in \mathcal{D}_{p}}2L(a-b)(a-b)^{\mathsf{T}}\cdot \mathbb{I}(\|L(a-b)\|^{2}\le 1)
\end{array}
\end{equation}
where $\mathbb{I}(\cdot)$ is the indicator function.

\begin{figure}[t]
{\scriptsize
\begin{Verbatim}[frame=single]
// Data-Parallel Distance Metric Learning
\end{Verbatim}
\vspace{-0.7cm}
\begin{Verbatim}[frame=single]
schedule() { // Empty, do nothing }
\end{Verbatim}
\vspace{-0.7cm}
\begin{Verbatim}[frame=single]
push() {
  L_local = PS.get(L) // Bounded-async read from param server
  change = 0
  for c=1..C    // Minibatch size C
    (x,y) = draw_similar_pair(DATA)
    (a,b) = draw_dissimilar_pair(DATA)
    change += DeltaL(L_local,x,y,a,b)  // SGD from Eq 7
  PS.inc(L,change/C)  // Add gradient to param server
}
\end{Verbatim}
\vspace{-0.7cm}
\begin{Verbatim}[frame=single]
pull() { // Empty, do nothing }
\end{Verbatim}
}
\vspace{-0.7cm}
\caption{\small Petuum DML data-parallel pseudocode.}
\label{fig:dml_alg}
\vspace{-0.5cm}
\end{figure}

Figure~\ref{fig:dml_alg} shows pseudocode for Petuum DML, which is simple to implement because the parameter server system {\tt PS} abstracts away complex networking code under a simple {\tt get()/read()} API. Moreover, the {\tt PS} automatically ensures high-throughput execution, via a bounded-asynchronous consistency model (Stale Synchronous Parallel) that can provide workers with stale local copies of the parameters $L$, instead of forcing workers to wait for network communication. Later, we will review the strong consistency and convergence guarantees provided by the SSP model.

Since DML is a data-parallel algorithm, only the parallel update {\tt push()} needs to be implemented (Figure~\ref{fig:dml_alg}). The scheduling function {\tt schedule()} is empty (because every worker touches every model parameter $L$), and we do not need aggregation {\tt push()} for this SGD algorithm. In our next example, we will show how {\tt schedule()} and {\tt push()} can be used to implement model-parallel execution.

\vspace{-0.2cm}
\subsection{Model-Parallel Lasso}

Lasso 
is a widely  used model to select features in high-dimensional problems, such as gene-disease association studies, or in online advertising via $\ell_1$-penalized regression~\cite{trench}.
Lasso takes the form of an optimization problem: 
\begin{align}
\min_{\bbeta} \ell(\bX, \by, \bbeta) + \lambda \sum_{j} |\beta_j|,
\label{eq:lasso}
\end{align}
where $\lambda$ denotes a regularization parameter that determines the sparsity of $\bbeta$, and $\ell(\cdot)$ is a non-negative convex loss function such as squared-loss or logistic-loss;  we assume that $\bX$ and $\by$ are standardized and consider \eqref{eq:lasso} without an intercept. For simplicity but without loss of generality, we let  $\ell(\bX,\by,\bbeta) = \frac{1}{2}\left\| \by - \bX \bbeta \right\|_2^2$; other loss functions (e.g. logistic) are straightforward and can be solved using the same approach~\cite{shotgun}.
We shall solve this via a coordinate descent (CD) model-parallel approach, similar but not identical to~\cite{shotgun,feature_cluster}.


\begin{figure}[t]
{\scriptsize
\begin{Verbatim}[frame=single]
// Model-Parallel Lasso
\end{Verbatim}
\vspace{-0.7cm}
\begin{Verbatim}[frame=single]
schedule() {
  for j=1..J     // Update priorities for all coeffs beta_j
    c_j = square(beta_j) + eta // Magnitude prioritization
  (s_1, ..., s_L') = random_draw(distribution(c_1, ..., c_J))
  // Choose L<L' pairwise-independent beta_j
  (j_1, ..., j_L) = correlation_check(s_1, ..., s_L')
  return (j_1, ..., j_L)
}
\end{Verbatim}
\vspace{-0.7cm}
\begin{Verbatim}[frame=single]
push(p = worker_id(), (j_1, ..., j_L) = schedule() ) {
  // Partial computation for L chosen beta_j; calls PS.get(beta)    
  (z_p[j_1], ..., z_p[j_L]) = partial(DATA[p], j_1, ..., j_L)  
  return z_p
}
\end{Verbatim}
\vspace{-0.7cm}
\begin{Verbatim}[frame=single]
pull((j_1, ..., j_L) = schedule(),
     (z_1, ..., z_P) = (push(1), ..., push(P)) ) {
  for a=1..L    // Aggregate partial computation from P workers
    newval = sum_threshold(z_1[j_a], ..., z_P[j_a])
    PS.put(beta[j_a], newval)  // Overwrite to parameter server
}
\end{Verbatim}
}
\vspace{-0.7cm}
\caption{\small Petuum Lasso model-parallel  pseudocode.}
\label{fig:lasso_alg}
\vspace{-0.5cm}
\end{figure}

The simplest parallel CD Lasso , shotgun~\cite{shotgun}, selects a random subset of parameters to be updated in parallel.
We now present a scheduled model-parallel Lasso that improves upon shotgun:
the Petuum scheduler chooses parameters that are nearly independent with each other, thus guaranteeing convergence of the Lasso objective. In addition, it prioritizes these parameters based on their distance to convergence, thus speeding up optimization.

Why is it important to choose independent parameters via scheduling? Parameter dependencies affect the CD update equation in the following manner: by taking the gradient of \eqref{eq:lasso}, we obtain the CD 
update for $\beta_j$:
\begin{align}
\beta_j^{(t)} \leftarrow S(
\bx_j^T\by 
-\sum_{k \neq j} \bx_j^T \bx_k \beta_k^{(t-1)},\lambda),
\label{eq:update_rule}
\end{align}
where $S(\cdot,\lambda)$ is a soft-thresholding operator, 
defined by $S(\beta_j,\lambda) \equiv \mbox{sign}(\beta)\left( \left| \beta \right| - \lambda \right)$.
In \eqref{eq:update_rule}, if $\bx_j^T \bx_k \neq 0$ (i.e., nonzero correlation) and
$\beta_j^{(t-1)} \neq 0$ and $\beta_k^{(t-1)} \neq 0$, then
a coupling effect is created between the two features $\beta_j$ and $\beta_k$. Hence, they are no longer
conditionally independent given the data: $\beta_j \not\perp \beta_k | \bX, \by$.
If the $j$-th and the $k$-th coefficients are updated concurrently, 
parallelization error may occur, causing the Lasso problem to converge slowly (or even diverge outright).

Petuum's  {\tt schedule()},  {\tt push()} and  {\tt pull()} interface is readily suited to implementing
scheduled model-parallel Lasso.
We use {\tt schedule()} to choose parameters
with low dependency, and to prioritize non-converged parameters. 
Petuum pipelines  {\tt schedule()} and {\tt push()}; thus
{\tt schedule()} does not slow down workers running {\tt push()}.
Furthermore, by separating the scheduling code {\tt schedule()} from the core optimization code {\tt push()} and {\tt pull()}, Petuum makes it easy to experiment
with complex scheduling policies that involve prioritization and dependency checking, thus facilitating the implementation
of new model-parallel algorithms --- for example, one could use {\tt schedule()} to prioritize according to KKT conditions in a constrained optimization problem, or to perform graph-based dependency checking like in Graphlab~\cite{graphlab12}. 
Later, we will show that the above Lasso schedule {\tt schedule()} is guaranteed to converge, and gives us near optimal solutions by controlling errors from parallel execution. The pseudocode for scheduled model parallel Lasso under Petuum  is shown in Figure \ref{fig:lasso_alg}.

\vspace{-0.2cm}
\subsection{Other Algorithms}

We have implemented other data- and model-parallel algorithms on Petuum as well.
Here, we briefly mention a few, while noting that many others are included in the Petuum open-source library.


\textbf{Topic Model (LDA):}
For LDA, the key parameter is the ``word-topic" table, that needs to be updated by all worker machines. We adopt a simultaneous data-and-model-parallel approach to LDA, and use a fixed schedule function {\tt schedule\_fix()} to cycle disjoint subsets of the word-topic table and data across machines for updating (via {\tt push()} and {\tt pull()}), without violating structural dependencies in LDA.

\textbf{Matrix Factorization (MF):}
High-rank decompositions of large matrices for improved accuracy~\cite{large_netflix08} can be solved by a model-parallel approach, and we implement it via a fixed schedule function {\tt schedule\_fix()}, where each worker machine only performs the model update {\tt push()} on a disjoint, unchanging subset of factor matrix rows.

\textbf{Deep Learning (DL):}
We implemented two types on Petuum: a general-purpose fully-connected Deep Neural Network (DNN) using the cross-entropy loss, and a Convolutional Neural Network (CNN) for image classification based off the open-source Caffe project. 
We adopt a data-parallel strategy {\tt schedule\_fix()}, where each worker uses its data subset to perform updates {\tt push()} to the full model $A$. While this data-parallel strategy could be amenable to MapReduce, Spark and GraphLab, we are not aware of DL implementations on those platforms.

\vspace{-0.3cm}
\section{Principles and Theory}
\vspace{-0.1cm}

\begin{figure}[t]
\vspace{-0.4cm}
\centering
\includegraphics[width=0.8\linewidth]{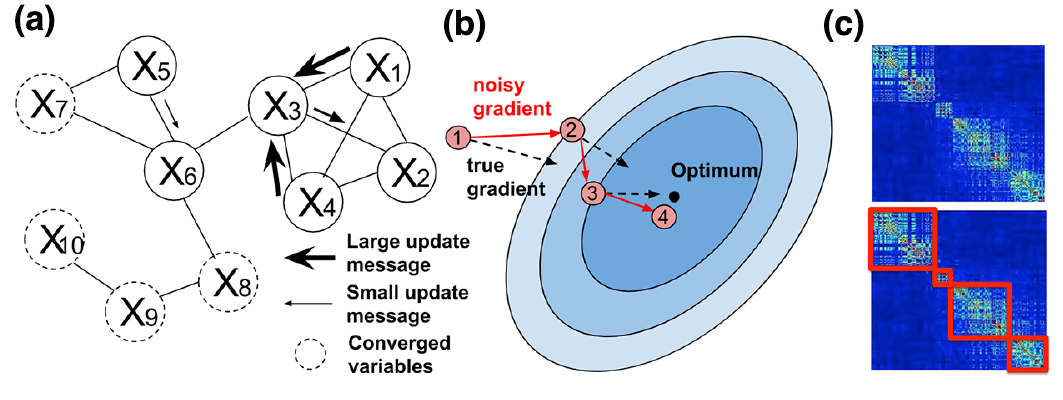}
\vspace{-0.6cm}
\caption{\small
Key properties of ML algorithms: (a) Non-uniform convergence; (b) Error-tolerant convergence; (c) Dependency structures amongst variables.
}
\label{fig:illustration}
\vspace{-0.6cm}
\end{figure}

Our iterative-convergent formulation of ML programs, and the explicit notion of data and model parallelism, make it convenient to explore three key properties of ML programs --- error-tolerant convergence, non-uniform convergence, dependency structures (Fig.~\ref{fig:illustration}) --- and to analyze how Petuum exploits these properties in a theoretically-sound manner to speed up ML program completion at Big Learning scales.

Some of these properties have previously been successfully exploited by a number of bespoke, large-scale implementations of popular ML algorithms: e.g. topic models~\cite{yuan2015lightlda,muli_osdi14}, matrix factorization~\cite{yu2012scalable,kumar2014fugue}, and deep learning~\cite{cat_recognizer12}. It is notable that MapReduce-style systems (such as Hadoop~\cite{white2012hadoop} and Spark~\cite{spark}) often do not fare competitively against these custom-built ML implementations, and one of the reasons is that these key ML properties are difficult to exploit under a MapReduce-like abstraction. Other abstractions may offer a limited degree of opportunity --- for example, vertex programming~\cite{graphlab12} permits graph dependencies to influence model-parallel execution.

\vspace{-0.2cm}
\subsection{Error tolerant convergence}
Data-parallel ML algorithms are often robust against minor errors in intermediate calculations; as a consequence, they still execute correctly even when their model parameters $A$ experience synchronization delays (i.e. the $P$ workers only see old or stale parameters), provided those delays are strictly bounded
\cite{hogwild, ho13, dai2015high, slow_learners, agarwal_delay, fugue}.
Petuum exploits this error-tolerance to reduce network communication/synchronization overheads substantially, by implementing the Stale Synchronous Parallel (SSP) consistency model~\cite{ho13,dai2015high} on top of the parameter server system, which provides all machines with access to the parameters $A$.

The SSP consistency model guarantees that if a worker reads from parameter server at iteration $c$, it is guaranteed to receive all updates from all workers computed at and before iteration $c - s - 1$, where $s$ is the staleness threshold.
If this is impossible because some straggling worker is more than $s$ iterations behind, the reader will stop until the straggler catches up and sends its updates. For stochastic gradient descent algorithms (such as the DML program), SSP has very attractive theoretical properties~\cite{dai2015high}, which we partially re-state here:\\[-0.6cm]

\begin{theorem}[adapted from \cite{dai2015high}]
{\bf SGD under SSP, convergence in probability: } 
Let $f(\bx)=\sum_{t=1}^T f_t(\bx)$ be a convex function, where the $f_t$ are also convex. We search for a minimizer $\bx^*$ via stochastic gradient descent on each component $\nabla f_t$ under SSP, with staleness parameter $s$ and $P$ workers. Let $\bu_t:=-\eta_t \nabla_t f_t(\tilde{\bx}_t)$ with $\eta_t=\frac{\eta}{\sqrt{t}}$. Under suitable conditions ($f_t$ are $L$-Lipschitz and bounded divergence $D(x||x')\le F^2$), we have
\begin{align*}
P\left[\frac{R\left[X\right]}{T} - \frac{1}{\sqrt{T}} \left(\eta L^{2} + \frac{F^{2}}{\eta} + 2\eta L^2\mu_{\gamma} \right) \ge \tau\right] 
\\
\le \exp\left\{\frac{-T\tau^2}{2\bar{\eta}_T\sigma_{\gamma} + \frac{2}{3}\eta L^2(2s+1)P\tau}\right\}
\end{align*}
where $R[X] := \sum_{t=1}^T f_t(\tilde{x}_t) - f(x^*)$, and $\bar{\eta}_T = \frac{\eta^2 L^4 (\ln T + 1)}{T} = o(T)$.
\end{theorem}
This means that $\frac{R[X]}{T}$ converges to $O(T^{-1/2})$ in probability with an exponential tail-bound; convergence is faster when the observed staleness average $\mu_{\gamma}$ and variance $\sigma_{\gamma}$ are smaller (and SSP ensures both $\mu_{\gamma},\sigma_{\gamma}$ are as small as possible). Dai {\it et al.} also showed that the variance of $\bx$ can be bounded, ensuring reliability and stability near an optimum~\cite{dai2015high}.


 

\vspace{-0.2cm}
\subsection{Dependency structures}
Naive parallelization of model-parallel algorithms (e.g. coordinate descent) may lead to uncontrolled parallelization error and non-convergence, caused by inter-parameter dependencies in the model. Such dependencies have been thoroughly analyzed under fixed execution schedules (where each worker updates the same set of parameters every iteration)
\cite{feature_cluster, shotgun,richtarik2012parallel},
but there has been little research on {\it dynamic schedules} that can react to changing model dependencies or model state $A$. Petuum's scheduler allows users to write dynamic scheduling functions $S_p^{(t)}(A^{(t)})$ --- whose output is a set of model indices $\{j_1,j_20,\dots\}$, telling worker $p$ to update $A_{j_1},A_{j_2},\dots$ --- as per their application's needs. This enables ML programs to analyze dependencies at run time (implemented via {\tt schedule()}), and select subsets of independent (or nearly-independent) parameters for parallel updates.

To motivate this, we consider a generic optimization problem, which many regularized regression problems --- including the Petuum Lasso example --- fit into:

\begin{align}
\min_{\wvec\in\RR^d} f(\wvec) + r(\wvec), \label{eq:generic_regression}
\end{align}
where $r(\wvec) = \sum_i r(w_i)$ is separable and $f$ has $\beta$-Lipschitz continuous gradient in the following sense:
\begin{align}
\label{eq:lcg}
f(\wvec + \zvec) \leq f(\wvec) + \zvec^\top \nabla f(\wvec) + \tfrac{\beta}{2}\zvec^\top X^\top X \zvec,
\end{align}
where $X = [\xvec_1, \ldots, \xvec_d]$ are $d$ feature vectors. W.l.o.g., we assume that each feature vector $\xvec_i$ is normalized, i.e., $\|\xvec_i\|_2  = 1, i = 1,\ldots, d$. Therefore $ |\xvec_i^\top \xvec_j| \leq 1$ for all $i, j$.

In the regression setting, $f(\wvec)$ represents a least-squares loss, $r(\wvec)$ represents a separable regularizer (e.g. $\ell_1$ penalty), and $\xvec_i$ represents the $i$-th feature column of the design (data) matrix, each element in $\xvec_i$ is a separate data sample. In particular, $|\xvec_i^\top \xvec_j|$ is the correlation between the $i$-th and $j$-th feature columns. The parameters $\wvec$ are simply the regression coefficients.

In the context of the model-parallel equation (\ref{eq:modelpara}), we can map the model $A = \wvec$, the data $D = X$, and the update equation $\Delta(A,S_p(A))$ to
\begin{align}
w_{j_p}^+ \gets \arg\min_{z\in \RR} \tfrac{\beta}{2}[z - (w_{j_p} - \tfrac{1}{\beta}g_{j_p})]^2 + r(z),
\end{align}
where $S_p^{(t)}(A)$ has selected a single coordinate $j_p$ to be updated by worker $p$ --- thus, $P$ coordinates are updated in every iteration. The aggregation function $F()$ simply allows each update $w_{j_p}$ to pass through without change.

The effectiveness of parallel coordinate descent depends on how the schedule $S_p^{(t)}()$ selects the coordinates $j_p$. In particular, naive random selection can lead to poor convergence rate or even divergence, with error proportional to the correlation $|\xvec_{j_a}^\top \xvec_{j_b}|$ between the randomly-selected coordinates $j_a,j_b$~\cite{feature_cluster,shotgun}. An effective and cheaply-computable schedule $S_{RRP,p}^{(t)}()$ 
involves randomly proposing a small set of $Q>P$ features $\{j_1,\dots,j_Q\}$, and then finding $P$ features in this set such that $|\xvec_{j_a}^\top \xvec_{j_b}|\le \theta$ for some threshold $\theta$, where $j_a,j_b$ are any two features in the set of $P$. This requires at most $\mathcal{O}(B^2)$ evaluations of $|\xvec_{j_a}^\top \xvec_{j_b}|\le \theta$ (if we cannot find $P$ features that meet the criteria, we simply reduce the degree of parallelism). We have the following convergence theorem:

\vspace{-0.3cm}
\begin{theorem}
{\bf $S_{RRP}()$ convergence: }
Let $k=1$ and $\epsilon := \frac{d(P-1)(\rho-1)}{B(B-1)\tau^2}  \approx \frac{(P-1)(\rho-1)}{d} < 1$, for constants $d,B$. After $t$ iterations,
\begin{align}
\EE[F(\wvec^{(t)}) - F(\wvec^\star)] \leq \frac{Cd\beta}{P(1-\epsilon)} \frac{1}{t},
\end{align}
where $F(\wvec) := f(\wvec) + r(\wvec)$ and $\wvec^\star$ is a minimizer of $F$.
\end{theorem}
For reference, the Petuum Lasso scheduler uses $S_{RRP}()$, augmented with a prioritizer we will describe soon.

In addition to asymptotic convergence, we show that $S_{RRP}$'s trajectory is close to ideal parallel execution:

\vspace{-0.3cm}
\begin{theorem}
{\bf $S_{RRP}()$ is close to ideal execution: }
Let $S_{ideal}()$ be an oracle schedule that always proposes $P$ random features with zero correlation. Let $\wvec_{ideal}^{(t)}$ be its parameter trajectory, and let $\wvec_{RRP}^{(t)}$ be the parameter trajectory of $S_{RRP}()$. Then,
\begin{align}
E[|\wvec_{ideal}^{(t)} &- \wvec_{RRP}^{(t)}|] \leq \frac{2JPm}{(T+1)^2\hat{P}} L^2 X^TXC,
\end{align}
for constants $C,m,L,\hat{P}$.
\end{theorem}
The proofs for both theorems can be found in the online supplement\footnote{\url{http://petuum.github.io/papers/kdd15_supp.pdf}}.

$S_{RRP}()$ is different from Scherrer {\it et al.}~\cite{feature_cluster}, who pre-cluster all $M$ features before starting coordinate descent, in order to find ``blocks" of nearly-independent parameters. In the Big Data and especially Big Model setting, feature clustering can be prohibitive --- fundamentally, it requires $\mathcal{O}(M^2)$ evaluations of $|\xvec_i^\top \xvec_j|$ for all $M^2$ feature combinations $(i,j)$, and although greedy clustering algorithms can mitigate this to some extent, feature clustering is still impractical when $M$ is very large, as seen in some regression problems~\cite{trench}. The proposed $S_{RRP}()$ only needs to evaluate a small number of $|\xvec_i^\top \xvec_j|$ every iteration, and we explain next, the random selection can be replaced with {\it prioritization} to exploit non-uniform convergence in ML problems.



\vspace{-0.2cm} 
\subsection{Non-uniform convergence}
In model-parallel ML programs, it has been empirically observed that some parameters $A_j$ can converge in much fewer/more updates than other parameters~\cite{lee2014model}.
For instance, this happens in Lasso regression because the model enforces sparsity, so most parameters remain at zero throughout the algorithm, with low probability of becoming non-zero again. Prioritizing Lasso parameters according to their magnitude greatly improves convergence per iteration, by avoiding frequent (and wasteful) updates to zero parameters~\cite{lee2014model}.

We call this {\it non-uniform ML convergence}, which can be exploited via a dynamic scheduling function $S_p^{(t)}(A^{(t)})$ whose output changes according to the iteration $t$ --- for instance, we can write a scheduler $S_{mag}()$ that proposes parameters with probability proportional to their current magnitude $(A_j^{(t)})^2$. This $S_{mag}()$ can be combined with the earlier dependency structure checking, leading to a {\it dependency-aware, prioritizing scheduler}. Unlike the dependency structure issue, prioritization has not received as much attention in the ML literature, though it has been used to speed up the PageRank algorithm, which is iterative-convergent~\cite{zhang2013priter}.
 

The prioritizing schedule $S_{mag}()$ can be analyzed in the context of the Lasso problem. First, we rewrite it by duplicating original $J$ features with opposite sign:
$
F(\bbeta) := \min_{\bbeta} \frac{1}{2}\left\| \by - \bX \bbeta \right\|_2^2 + \lambda \sum_{j=1}^{2J} \beta_j.
$
Here, $\bX$ contains $2J$ features and $\beta_j\geq 0$, for all $j=1, \ldots, 2J$. \\[-0.6cm]

\begin{theorem} [Adapted from ~\cite{lee2014model}]
{\bf Optimality of \\ Lasso priority scheduler: }
Suppose $\mathcal{B}$ 
is the set of indices of coefficients updated in parallel
at the $t$-th iteration, and
$\rho$ is sufficiently small constant such that 
$\rho \delta \beta_j^{(t)} \delta \beta_k^{(t)} \approx 0$, 
for all $j \neq k \in \mathcal{B}$.
Then, the sampling distribution
$p(j) \propto  (\delta \beta_j^{(t)} )^2$  
approximately maximizes a lower bound on 
$E_{\mathcal{B}} [F(\bbeta^{(t)})-F(\bbeta^{(t)}+ \Delta \bbeta^{(t)}) ]$. 
\end{theorem}
This theorem shows that a prioritizing scheduler
speeds up Lasso convergence by decreasing the objective as much as possible every iteration.
The pipelined Petuum scheduler system approximates
$p(j) \propto (\delta \beta_j^{(t)})^2$ with
$p'(j) \propto  \delta (\beta_j^{(t-1)})^2 + \eta$, because $\delta \beta_j^{(t)}$ is 
unavailable until all computations on $\beta_j^{(t)}$ have finished (and we want schedule before that happens, so that workers are fully occupied). Since we are approximating, we add a constant $\eta$ to ensure all $\beta_j$'s have a non-zero probability of being updated.

\vspace{-0.3cm}
\section{Performance}
\vspace{-0.1cm}


Petuum's ML-centric system design supports a variety of ML programs, and improves their performance on Big Data in the following senses:
(1) Petuum implementations of DML and Lasso achieve significantly faster convergence rate than baselines (i.e., DML implemented on single machine, and Shotgun \cite{shotgun});
(2) Petuum ML implementations can run faster than other platforms (e.g. Spark, GraphLab\footnote{We omit Hadoop, as it is well-established that Spark and GraphLab significantly outperform it~\cite{spark,graphlab12}.}), because Petuum can exploit model dependencies, uneven convergence and error tolerance; (3) Petuum ML implementations can reach larger model sizes than other platforms, because Petuum stores ML program variables in a lightweight fashion (on the parameter server and scheduler); (4) for ML programs without distributed implementations, we can implement them on Petuum and show good scaling with an increasing number of machines. We emphasize that Petuum is, for the moment, primarily about allowing ML practitioners to implement and experiment with new data/model-parallel ML algorithms on small-to-medium clusters; Petuum currently lacks features that are necessary for clusters with $\ge1000$ machines, such as automatic recovery from machine failure. Our experiments are therefore focused on clusters with 10-100 machines, in accordance with our target users.

\noindent{\bf Performance of Distance Metric Learning and Lasso}

\begin{figure}[t]
\vspace{-0.4cm}
\centering
\includegraphics[width=0.49\linewidth]{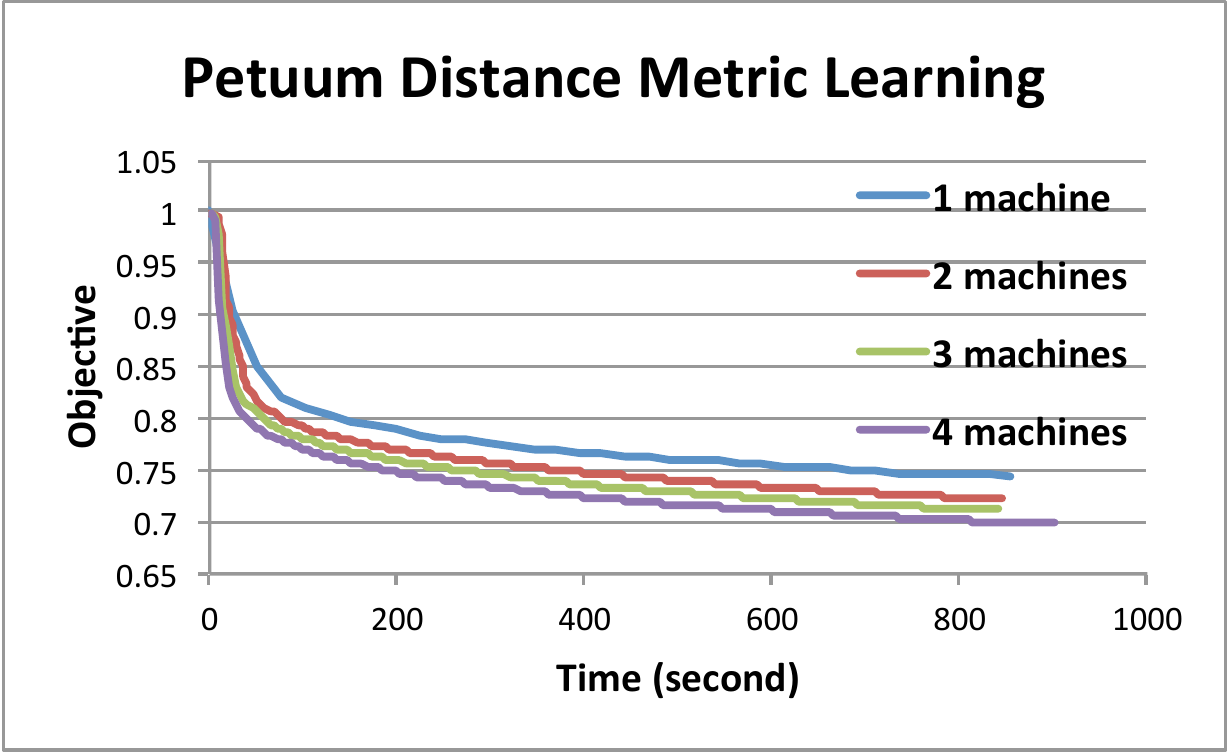}\nolinebreak 
\includegraphics[width=0.5\linewidth]{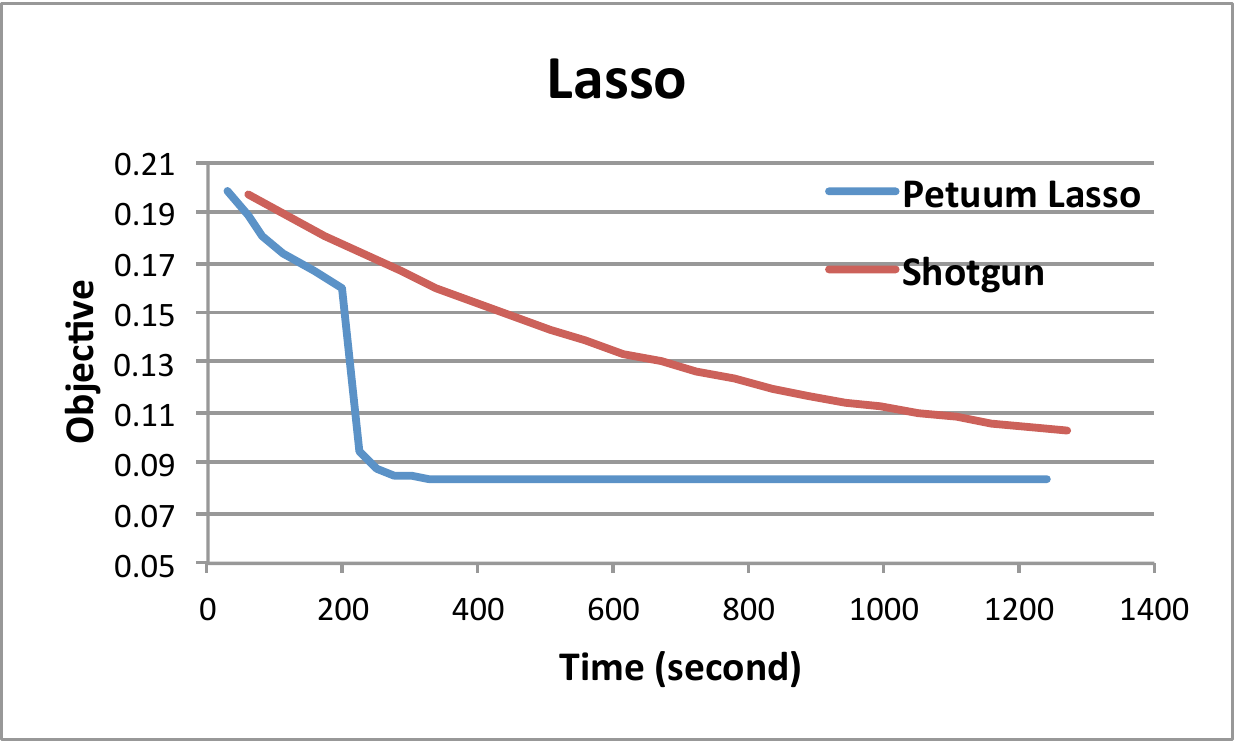}
\vspace{-0.6cm}
\caption{\small {\bf Left:} 
Petuum DML convergence curve 
with different number of machines from 1 to 4.
{\bf Right:} 
Lasso convergence curve
by Petumm Lasso and Shotgun. 
}
\label{fig:convergence_curves}
\vspace{-0.5cm}
\end{figure}

We first demonstrate the performance of DML and lasso,
implemented under Petuum.
In Figure \ref{fig:convergence_curves}, 
we showcase the convergence of Petuum and baselines using a fixed model size
(we used a $21504\times 1000$ distance matrix for DML; 100M features for Lasso).
For DML, increasing the number of machines
consistently increases the convergence speed. Petuum DML achieves 3.8 times speedup with 4 machines and 1.9 times speedup with 2 machines, demonstrating that Petuum DML has the potential to scale very well with more machines.
For Lasso, given the same number of machines,
Petuum achieved a significantly faster convergence rate
than Shotgun (which randomly selects
a subset of parameters to be updated). 
In the initial stage, Petuum lasso and Shotgun show similar convergence
rates because Petuum updates every parameter in the first iteration
to ``bootstrap" the scheduler (at least one iteration is required to initialize
all parameters). After this initial stage, Petuum dramatically 
decreases the Lasso objective compared to Shotgun, by taking advantage of 
dependency structures and non-uniform convergence via the scheduler.

\begin{figure}[t]
\vspace{-0.4cm}
\centering
\includegraphics[width=0.47\linewidth]{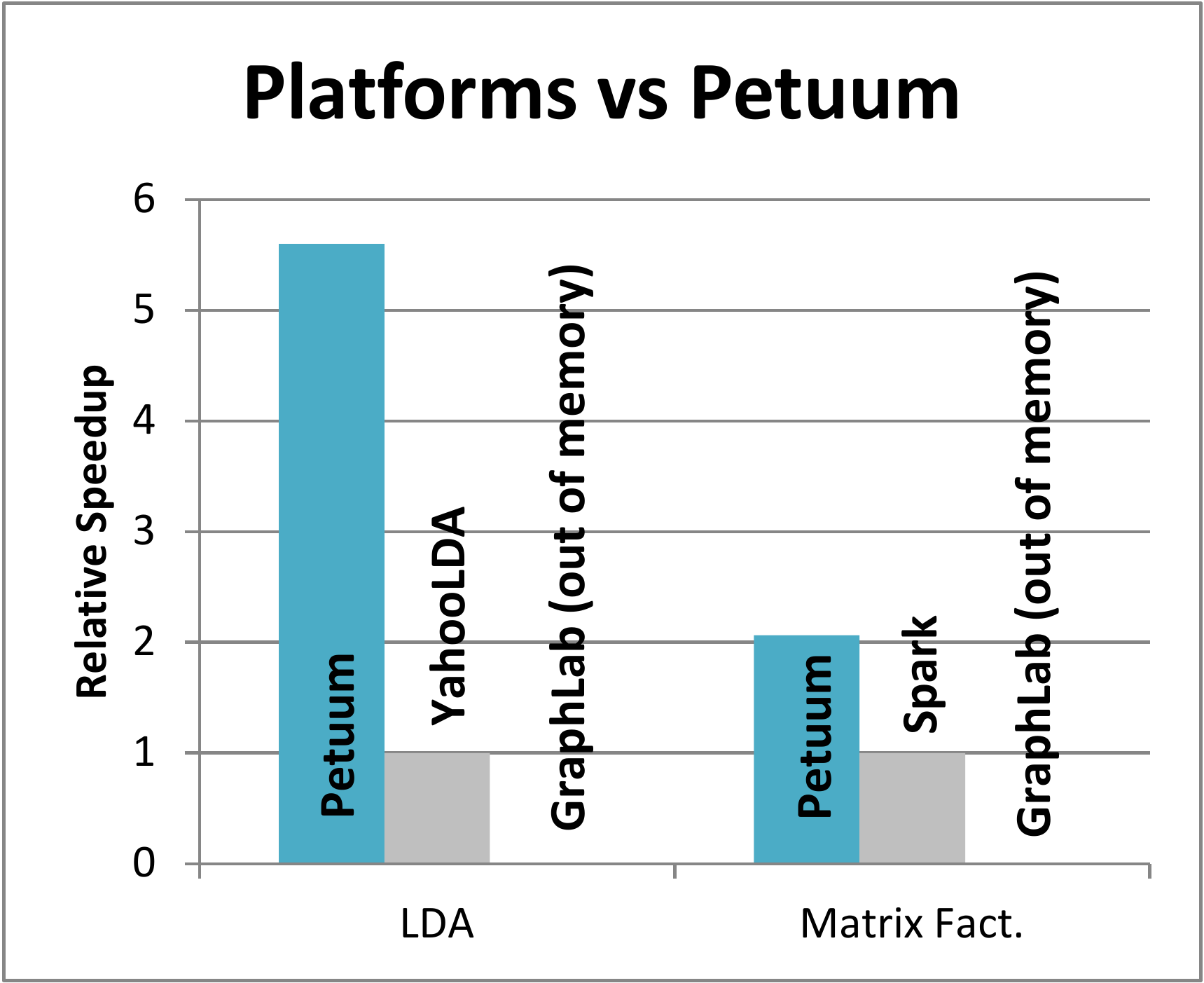}\nolinebreak
\includegraphics[width=0.51\linewidth]{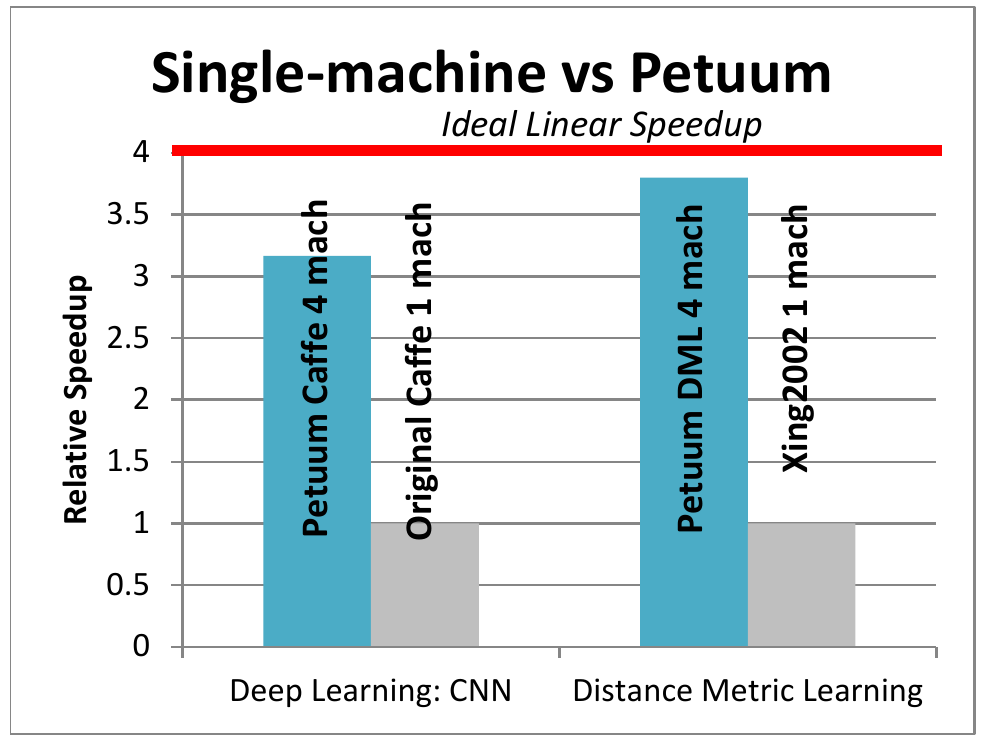} 
\vspace{-0.3cm}
\caption{\small 
{\bf Left:} Petuum performance: relative speedup vs popular platforms (larger is better). Across ML programs, Petuum is at least 2-10 times faster than popular implementations. {\bf Right:} Petuum is a good platform for writing cluster versions of existing single-machine algorithms, achieving near-linear speedup with increasing number of machines (Caffe CNN and DML). 
}
\label{fig:main_results}
\vspace{-0.4cm}
\end{figure}

\noindent{\bf Platform Comparison}
Figure \ref{fig:main_results} (left) compares Petuum to popular ML platforms (Spark and GraphLab) and well-known cluster implementations (YahooLDA). For two common ML programs of LDA and MF, we show the relative speedup of Petuum over the other platforms' implementations. 
In general, Petuum is between 2-6 times faster than other platforms; the differences help to illustrate the various data/model-parallel features in Petuum. For MF, Petuum uses the same model-parallel approach as Spark and GraphLab, but it performs twice as fast as Spark,
while GraphLab ran out of memory. On the other hand, Petuum LDA is nearly 6 times faster than YahooLDA;
the speedup mostly comes from scheduling $S()$, which enables correct, dependency-aware model-parallel execution. 

\noindent{\bf Scaling to Larger Models}

\begin{figure}[t]
\centering
\includegraphics[width=0.34\linewidth]{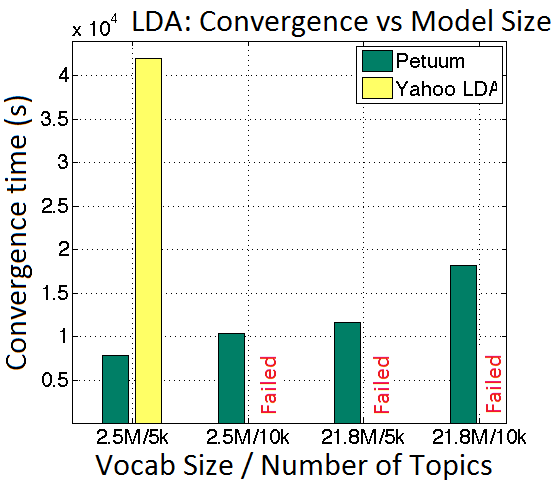}
\includegraphics[width=0.64\linewidth]{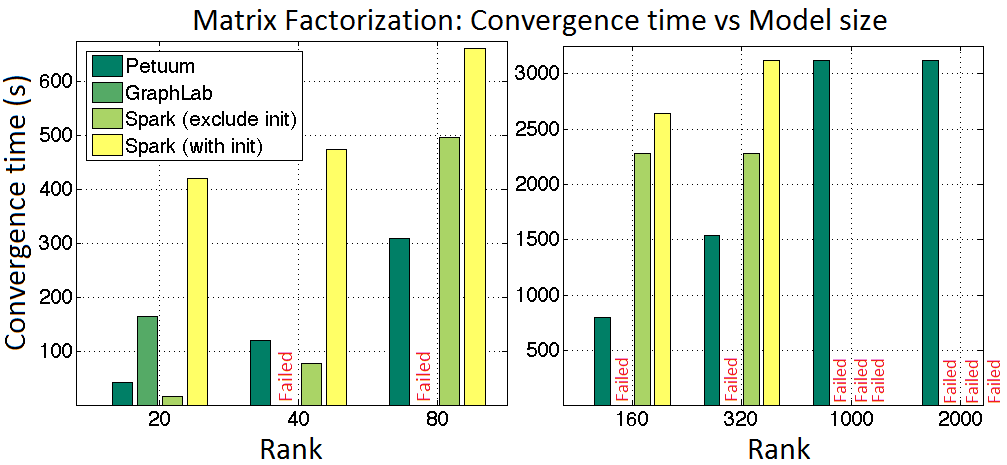}
\vspace{-0.6cm}
\caption{\small {\bf Left:} LDA convergence time: Petuum vs YahooLDA (lower is better). Petuum's data-and-model-parallel LDA converges faster than YahooLDA's data-parallel-only implementation, and scales to more LDA parameters (larger vocab size, number of topics). {\bf Right panels:} Matrix Factorization convergence time: Petuum vs GraphLab vs Spark. Petuum is fastest and the most memory-efficient, and is the only platform that could handle Big MF models with rank $K\ge 1000$ on the given hardware budget.}
\label{fig:modelsize_results}
\vspace{-0.6cm}
\end{figure}

Here, we show that Petuum supports larger ML models for the same amount of cluster memory. Figure \ref{fig:modelsize_results} shows ML program running time versus model size, given a fixed number of machines --- the left panel compares Petuum LDA and YahooLDA; PetuumLDA converges faster and supports LDA models that are $>10$ times larger\footnote{LDA model size is equal to vocab size times number of topics.}, allowing long-tail topics to be captured. The right panels compare Petuum MF versus Spark and GraphLab; again Petuum is faster and supports much larger MF models (higher rank) than either baseline. Petuum's model scalability is the result of two factors: (1) model-parallelism, which divides the model across machines; (2) a lightweight parameter server system with minimal storage overhead.

\noindent{\bf Fast Cluster Implementations of New ML Programs}

We show that Petuum facilitates the development of new ML programs without existing cluster implementations. In Figure \ref{fig:main_results} (right), we present two instances: first, a cluster version of the open-source Caffe CNN toolkit, created by adding $\sim600$ lines of Petuum code. The basic data-parallel strategy was left unchanged, so the Petuum port directly tests Petuum's efficiency. Compared to the original single-machine Caffe {\it with no network communication}, Petuum achieves approaching-linear speedup ($3.1$-times speedup on 4 machines) due to the parameter server's SSP consistency for managing network communication. Second, we compare the Petuum DML program against the original DML algorithm proposed in \cite{xing2002distance} (denoted by Xing2002), which is optimized with SGD on a single-machine (with parallelization over matrix operations). The intent is to show that a fairly simple data-parallel SGD implementation of DML (the Petuum program) can greatly speed up execution over a cluster. The Petuum implementation converges 3.8 times faster than Xing2002 on 4 machines
--- this provides evidence that Petuum enables data/model-parallel algorithms to be efficiently implemented over clusters.

\noindent{\bf Experimental settings}

We used 3 clusters with varying specifications, demonstrating Petuum's adaptability to different hardware: ``Cluster-1" has machines with 2 AMD cores, 8GB RAM, 1Gbps Ethernet; ``Cluster-2" has machines with 64 AMD cores, 128GB RAM, 40Gbps Infiniband; ``Cluster-3" has machines with 16 Intel cores, 128GB RAM, 10Gbps Ethernet.

LDA was run on 128 Cluster-1 nodes, using 3.9m English Wikipedia abstracts with unigram ($V=2.5$m) and bigram ($V=21.8$m) vocabularies. MF and Lasso were run on 10 Cluster-2 nodes, respectively using the Netflix data and a synthetic Lasso dataset with $N=50$k samples and 100m features/parameters. CNN was run on 4 Cluster-3 nodes, using a 250k subset of Imagenet with 200 classes, and 1.3m model parameters.
The DML experiment was run on 4 Cluster-2 nodes, using the 1-million-sample Imagenet \cite{deng2009imagenet} dataset with 1000 classes (220m model parameters), and 200m similar/dissimilar statements. 

\vspace{-0.3cm}
\small{
\bibliographystyle{abbrv}
\bibliography{literature}
}

\newpage\appendix

\section{Proof of Theorem 2}

We prove that the Petuum $S_{RRP}()$ scheduler makes the Regularized Regression Problem converge. We note that $S_{RRP}()$ has the following properties: (1) the scheduler \emph{uniformly} randomly selects $Q$ out of $d$ coordinates (where $d$ is the number of features); (2) the scheduler performs {\it dependency checking} and retains $P$ out of $Q$ coordinates; (3) in parallel, each of the $P$ workers is assigned one coordinate, and performs coordinate descent on it:
\begin{align}
\label{eq:pg}
w_{j_p}^+ \gets \arg\min_{z\in \RR} \tfrac{\beta}{2}[z - (w_{j_p} - \tfrac{1}{\beta}g_{j_p})]^2 + r(z),
\end{align}
where $g_j = \nabla_j f(\wvec)$ is the $j$-th partial derivative, and the coordinate $j_p$ is assigned to the $p$-th worker.
Note that \eqref{eq:pg} is simply the gradient update: $w \gets w - \tfrac{1}{\beta}g$,
followed by applying the proximity operator of $r$.

One way for the scheduler to select $P$ coordinates into $Q$ is to perform {\it dependency checking}: Coordinates $i$ and $j$ are in the same block iff $|\xvec_i^\top \xvec_j| \leq \theta$ for some parameter $\theta \in (0,1)$. Consider the following matrix 
\begin{align}
\forall i, A_{ii} = 1, \quad \forall i\ne j,
A_{ij} = \begin{cases}
\xvec_i^\top \xvec_j, & \mbox{ if } |\xvec_i^\top \xvec_j| \leq \theta \\
0, & \mbox{ otherwise }
\end{cases}
,
\end{align}
whose spectral radius $\rho = \rho(A)$ will play a major role in our analysis. A trivial bound for the spectral radius $\rho(A)$ is:
\begin{align}
|\rho - 1| \leq \sum_{j\ne i} |A_{ij}| \leq (d-1)\theta.
\end{align}
Thus, if $\theta$ is small, the spectral radius $\rho$ is small.

Denote $N$ the total number of pairs $(i,j)$ that can pass the dependency check. Roughly $N \sim O(d^2)$ if $\theta$ is close to 1 (i.e., all possible pairs). We assume that each of such pair will be selected by the scheduler with equal probability (i.e., $1/N$). This can be achieved by rejection sampling.
As a consequence, $P$, the number of coordinates selected by the scheduler, is a random variable and may vary from step to step. In practice, we assume that $P$ is equal to the number of available workers.

\paragraph{Theorem 2}
Let $\epsilon := \frac{d(\EE P^2/\EE P-1)(\rho-1)}{N} \approx \frac{(\EE P-1)(\rho-1)}{d} < 1$, then after $t$ steps, we have
\begin{align}
\EE[F(\wvec_t) - F(\wvec^\star)] \leq \frac{Cd\beta}{\EE P(1-\epsilon)} \frac{1}{t},
\end{align}
where $F(\wvec) := f(\wvec) + r(\wvec)$ and $\wvec^\star$ denotes a (global) minimizer of $F$ (whose existence is assumed for simplicity).

\paragraph{Proof of Theorem 2}
We first bound the algorithm's progress at step $t$. To avoid cumbersome double indices, let $\wvec = \wvec_t$ and $\zvec = \wvec_{t+1}$, then applying \eqref{eq:lcg}
\begin{align*}
&\EE [F(\zvec) - F(\wvec)] \\
&\leq \EE\bigg[\sum_{p=1}^P g_{j_p} (w_{j_p}^+ - w_{j_p}) + r(w_{j_p}^+) - r(w_{j_p}) \\
& + \frac{\beta}{2} (w_{j_p}^+ - w_{j_p})^2 + ~\frac{\beta}{2}\sum_{p\ne q} (w_{j_p}^+ - w_{j_p}) (w_{j_q}^+ - w_{j_q}) \xvec_{j_p}^\top \xvec_{j_q}\bigg]\\
&= \frac{\EE P}{d} \bigg[ \gvec^\top (\wvec^+ - \wvec)  + r(\wvec^+) - r(\wvec) + \frac{\beta}{2}\|\wvec^+-\wvec\|_2^2\bigg]\\
& \qquad \qquad + ~\frac{\beta \EE[P(P-1)]}{2N}(\wvec^+ - \wvec)^\top (A - I) (\wvec^+ - \wvec) \\
&\leq -\frac{\beta \EE P}{2d} \|\wvec^+ - \wvec\|_2^2 + \frac{\beta \EE[P(P - 1)](\rho-1)}{2N}\|\wvec^+ - \wvec\|_2^2 \\
&\leq -\frac{\beta \EE P (1-\epsilon)}{2d}\|\wvec^+ - \wvec\|_2^2,
\end{align*}
where we define $\epsilon = \frac{d(\EE P^2 / \EE P -1)(\rho-1)}{N}$, and the second inequality follows from the optimality of $\wvec^+$ as defined in \eqref{eq:pg}. Therefore as long as $\epsilon < 1$, the algorithm is decreasing the objective. This in turn puts a limit on the expected number of parallel workers $P$, roughly inverse proportional to the spectral radius $\rho$.

The rest of the proof follows the same line as that of shotgun \cite{shotgun}. To give a quick idea, consider the case where $0 \in \partial r(\wvec_t)$, then
$$
F(\wvec_{t+1}) - F(\wvec^\star) \leq (\wvec_{t+1} - \wvec^\star)^\top \gvec \leq \|\wvec_{t+1} - \wvec^\star\|_2 \cdot \|\gvec\|_2,
$$
and $\|\wvec_{t+1} - \wvec_{t}\|_2^2 = \|\gvec\|_2^2 / \beta^2$. Thus, defining $\delta_t = F(\wvec_t) - F(\wvec^\star)$, we have
\begin{align}
\EE(\delta_{t+1} - \delta_t)
&\leq -\frac{\EE P(1-\epsilon)}{2d\beta\|\wvec_{t+1} - \wvec^\star\|_2^2} \EE(\delta_{t+1}^2) \\
&\leq -\frac{\EE P(1-\epsilon)}{2d\beta\|\wvec_{t+1} - \wvec^\star\|_2^2} [\EE(\delta_{t+1})]^2.
\end{align}
Using induction it follows that $E(\delta_{t}) \leq \frac{Cd\beta}{\EE P(1-\epsilon)} \frac{1}{t}$ for some universal constant $C$. $\square$

The theorem confirms some intuition: The bigger the expected number of selected coordinates $\EE P$, the faster algorithm converges, but it also increases $\epsilon$, demonstrating a tradeoff among parallelization and correctness. The variance $\EE P^2$ also plays a role here: the smaller it is, the faster the algorithm converges (since $\epsilon$ is proportional to it). Of course, the bigger $N$ is, i.e., less coordinates are correlated above $\theta$, the faster the algorithm converges (since $\epsilon$ is inverse proportional to it).

\textbf{Remark:} We compare Theorem 2 with Shotgun \cite{shotgun} and the Block greedy algorithm in \cite{feature_cluster}. The convergence rate we get is similar to shotgun, but with a significant difference: Our spectral radius $\rho = \rho(A)$ is potentially much smaller than shotgun's $\rho(X^\top X)$, since by partitioning we zero out all entries in the correlation matrix $X^\top X$ that are bigger than the threshold $\theta$. In other words, we get to control the spectral radius while shotgun is totally passive.

The convergence rate in \cite{feature_cluster} is $\frac{CB}{P(1-\epsilon')}\frac{1}{t}$, where $\epsilon' = \frac{(P-1)(\rho'-1)}{B-1}$. Compared with ours, we have a bigger (hence worse) numerator ($d$ vs. $B$) but the denominator ($\epsilon'$ vs. $\epsilon$) are not directly comparable: we have a bigger spectral radius $\rho$ and bigger $d$ while \cite{feature_cluster} has a smaller spectral radius $\rho'$ (essentially taking a submatrix of our $A$) and smaller $B-1$. Nevertheless, we note that \cite{feature_cluster} may have a higher per-step complexity: each worker needs to check all of its assigned $\tau$ coordinates just to update one ``optimal'' coordinate. In contrast, we simply pick a random coordinate, and hence can be much cheaper per-step.

\section{Proof of Theorem 3}

For the Regularized Regression Problem, we prove that the Petuum $S_{RRP}()$ scheduler produces a solution trajectory $\wvec_{RRP}^{(t)}$ that is close to ideal execution:

\paragraph{Theorem 3}
($S_{RRP}()$ is close to ideal execution)
Let $S_{ideal}()$ be an oracle schedule that always proposes $P$ random features with zero correlation. Let $\wvec_{ideal}^{(t)}$ be its parameter trajectory, and let $\wvec_{RRP}^{(t)}$ be the parameter trajectory of $S_{RRP}()$. Then,
\begin{align}
E[|\wvec_{ideal}^{(t)} &- \wvec_{RRP}^{(t)}|] \leq \frac{2JPm}{(T+1)^2\hat{P}} L^2 X^TXC,
\end{align}
$C$ is a data dependent constant, $m$ is the strong convexity constant, $L$ is the domain width of $A_j$, and $\hat{P}$ is the expected number of indexes that $S_{RRP}()$ can actually parallelize in each iteration (since it may not be possible to find $P$ nearly-independent parameters).


We assume that the objective function $F(\wvec) = f(\wvec) + r(\wvec)$ is strongly convex --- for certain problems, this can be achieved through parameter replication, e.g. $\min_\wvec \frac{1}{2} ||y-X\wvec||^2_2 + \lambda\sum_{j=1}^{2M}\wvec_j$ is the replicated form of Lasso regression seen in Shotgun~\cite{shotgun}.


\paragraph{Lemma 1}  The difference between successive updates is:
\begin{equation}
F(\wvec+\Delta \wvec) - F(\wvec) \leq -(\Delta \wvec)^T \Delta \wvec + \frac{1}{2}(\Delta \wvec)^T X^TX \Delta \wvec
\end{equation}
\textbf{Proof:} The Taylor expansion of $F(\wvec+\Delta \wvec)$ around $\wvec$ coupled with the fact that $F(\wvec)^{'''}$ (3rd-order) and higher order derivatives are zero leads to the above result. $\square$

\paragraph{Proof of Theorem 3}
By using Lemma 1, and telescoping sum:
\begin{align}
&F(\wvec^{(T)}_{ideal}) - F(\wvec^{(0)}_{ideal}) = \nonumber\\
&\sum_{t=1}^{T} -(\Delta \wvec^{(t)}_{ideal})^\top \Delta \wvec^{(t)}_{ideal} + \frac{1}{2}(\Delta \wvec^{(t)}_{ideal})^\top X^\top X \Delta \wvec^{(t)}_{ideal} 
\end{align}
Since $S_{ideal}$ chooses $P$ features with 0 correlation, 
\begin{equation*}
F(\wvec^{(T)}_{ideal}) - F(\wvec^{(0)}_{ideal}) = \sum_{t=1}^{T} -(\Delta \wvec^{(t)}_{ideal})^\top \Delta \wvec^{(t)}_{ideal}
\end{equation*}
Again using Lemma 1, and telescoping sum:
\begin{align}
&F(\wvec^{(T)}_{RRP}) - F(\wvec^{(0)}_{RRP}) = \nonumber\\
&\sum_{t=1}^{T} -(\Delta \wvec^{(t)}_{RRP})^\top \Delta \wvec^{(t)}_{RRP} + \frac{1}{2}(\Delta \wvec^{(t)}_{RRP})^\top X^\top X \Delta \wvec^{(t)}_{RRP}
\end{align}
Taking the difference of the two sequences, we have:
\begin{align}
&F(\wvec^{(T)}_{ideal}) - F(\wvec^{(T)}_{RRP})  = \nonumber \\
&\left(\sum_{t=1}^{T} -(\Delta \wvec^{(t)}_{ideal})^\top \Delta \wvec^{(t)}_{ideal}\right) \nonumber \\
&- \left(\sum_{t=1}^{T} -(\Delta \wvec^{(t)}_{RRP})^\top \Delta \wvec^{(t)}_{RRP} + \frac{1}{2}(\Delta \wvec^{(t)}_{RRP})^\top X^\top X \Delta \wvec^{(t)}_{RRP}\right)
\end{align}
Taking expectations w.r.t. the randomness in iteration, indices chosen at each iteration, and the inherent randomness in the two sequences, we have:
\begin{align}
&E[|F(\wvec^{(T)}_{ideal}) - F(\wvec^{(T)}_{RRP})|]  = \nonumber \\
&E[|(\sum_{t=1}^{T} -(\Delta \wvec^{(t)}_{ideal})^T \Delta \wvec^{(t)}_{ideal}) \nonumber \\
&- (\sum_{t=1}^{T} -(\Delta \wvec^{(t)}_{RRP})^T \Delta \wvec^{(t)}_{RRP} + \frac{1}{2}(\Delta \wvec^{(t)}_{RRP})^\top X^\top X \Delta \wvec^{(t)}_{RRP})|] \nonumber \\
&= (C_{data}+\frac{1}{2})E[|\sum_{t=1}^{T}(\Delta \wvec^{(t)}_{RRP})^\top X^\top X \Delta \wvec^{(t)}_{RRP})|],
\end{align}
where $C_{data}$ is a data dependent constant. Here, the difference between 
$(\Delta \wvec^{(t)}_{ideal})^\top \Delta \wvec^{(t)}_{ideal}$ and $(\Delta \wvec^{(t)}_{RRP})^\top \Delta \wvec^{(t)}_{RRP}$ can only be possible due to $(\Delta \wvec^{(t)}_{RRP})^\top X^\top X \Delta \wvec^{(t)}_{RRP}$.

Following the proof in the shotgun paper~\cite{shotgun}, we get
\begin{align}
E[|F(\wvec^{(T)}_{ideal}) &- F(\wvec^{(T)}_{RRP})|] \leq \frac{2JP}{(T+1)^2\hat{P}} L^2 X^TXC,
\end{align}
where $C$ is a data dependent constant, $L$ is the domain width of $\wvec_j$ (i.e. the difference between its maximum and minimum possible values), and $\hat{P}$ is the expected number of indexes that $S_{RRP}()$ can actually parallelize in each iteration.

Finally, we apply the strong convexity assumption to get
\begin{align}
E[|\wvec^{(T)}_{ideal} &- \wvec^{(T)}_{RRP}|] \leq \frac{2JPm}{(T+1)^2\hat{P}} L^2 X^TXC,
\end{align}
where $m$ is the strong convexity constant. $\square$

\end{document}